\documentclass[10pt,twocolumn,letterpaper]{article}

\usepackage{cvpr}              

\usepackage{graphicx}
\usepackage{amsmath}
\usepackage{amssymb}
\usepackage{booktabs}
\usepackage{makecell}
\usepackage{CJK}
\usepackage{xcolor}
\usepackage{paralist}

\usepackage{colortbl}
\definecolor{light-gray}{gray}{0.92}

\usepackage[accsupp]{axessibility}  

\usepackage[pagebackref,breaklinks,colorlinks]{hyperref}

\usepackage[capitalize]{cleveref}
\crefname{section}{Sec.}{Secs.}
\Crefname{section}{Section}{Sections}
\Crefname{table}{Table}{Tables}
\crefname{table}{Tab.}{Tabs.}

\usepackage{amsthm}
\newtheorem{theorem}{Theorem}
\newtheorem{lemma}{Lemma}
\newtheorem{proposition}{Proposition}
\usepackage{mathtools}
\usepackage{cleveref}
\def\cvprPaperID{6986} 
\def\confName{CVPR}
\def\confYear{2023}

\begin{document}

\title{DeepMAD: Mathematical Architecture Design for\\ Deep Convolutional Neural Network}

\author{
Xuan Shen$^1$\thanks{These Authors contributed equally.}, Yaohua Wang$^2$\text{$^*$}, Ming Lin$^3$\thanks{Work done before joining Amazon.}, Yilun Huang$^2$, Hao Tang$^4$, Xiuyu Sun$^2$$^{\ddag}$, Yanzhi Wang$^1$\thanks{Corresponding Author}\\
$^1$Northeastern University, $^2$Alibaba Group, $^3$Amazon, $^4$ETH Zurich\\
{
\tt\small \{shen.xu,yanz.wang\}@northeastern.edu, \{xiachen.wyh,lielin.hyl,xiuyu.sxy\}@alibaba-inc.com,
}
\\
{\tt\small minglamz@amazon.com, hao.tang@vision.ee.ethz.ch}
}
\maketitle

\newif\ifmodify
\ifmodify

\newcommand{\cross}[1]{\textcolor{gray}{\sout{#1}}}
\newcommand{\todo}[1]{\textcolor{red}{#1}}

\newcommand{\xuan}[1]{\textcolor{blue}{#1}}
\newcommand{\xiachen}[1]{\textcolor{orange}{#1}}

\else

\newcommand{\cross}[1]{}
\newcommand{\todo}[1]{#1}

\newcommand{\xuan}[1]{#1}
\newcommand{\xiachen}[1]{#1}

\fi

\begin{abstract}

The rapid advances in Vision Transformer (ViT) refresh the state-of-the-art performances in various vision tasks, overshadowing the conventional CNN-based models. This ignites a few recent striking-back research in the CNN world showing that pure CNN models can achieve as good performance as ViT models when carefully tuned. While encouraging, designing such high-performance CNN models is challenging, requiring non-trivial prior knowledge of network design. To this end, a novel framework termed Mathematical Architecture Design for Deep CNN~(DeepMAD\footnote{Source codes are available at \url{https://github.com/alibaba/lightweight-neural-architecture-search}}) is proposed to design high-performance CNN models in a principled way. In DeepMAD, a CNN network is modeled as an information processing system whose expressiveness and effectiveness can be analytically formulated by their structural parameters. Then a constrained mathematical programming (MP) problem is proposed to optimize these structural parameters. The MP problem can be easily solved by off-the-shelf MP solvers on CPUs with a small memory footprint. In addition, DeepMAD is a pure mathematical framework: no GPU or training data is required during network design. The superiority of DeepMAD is validated on multiple large-scale computer vision benchmark datasets. 
Notably on ImageNet-1k, only using conventional convolutional layers, DeepMAD achieves 0.7\% and 1.5\% higher top-1 accuracy than ConvNeXt and Swin on Tiny level, and 0.8\% and 0.9\% higher on Small level.


\end{abstract}


\section{Introduction}
\label{sec:intro}




\begin{figure}[t]
  \centering
  \includegraphics[width=0.92\linewidth]{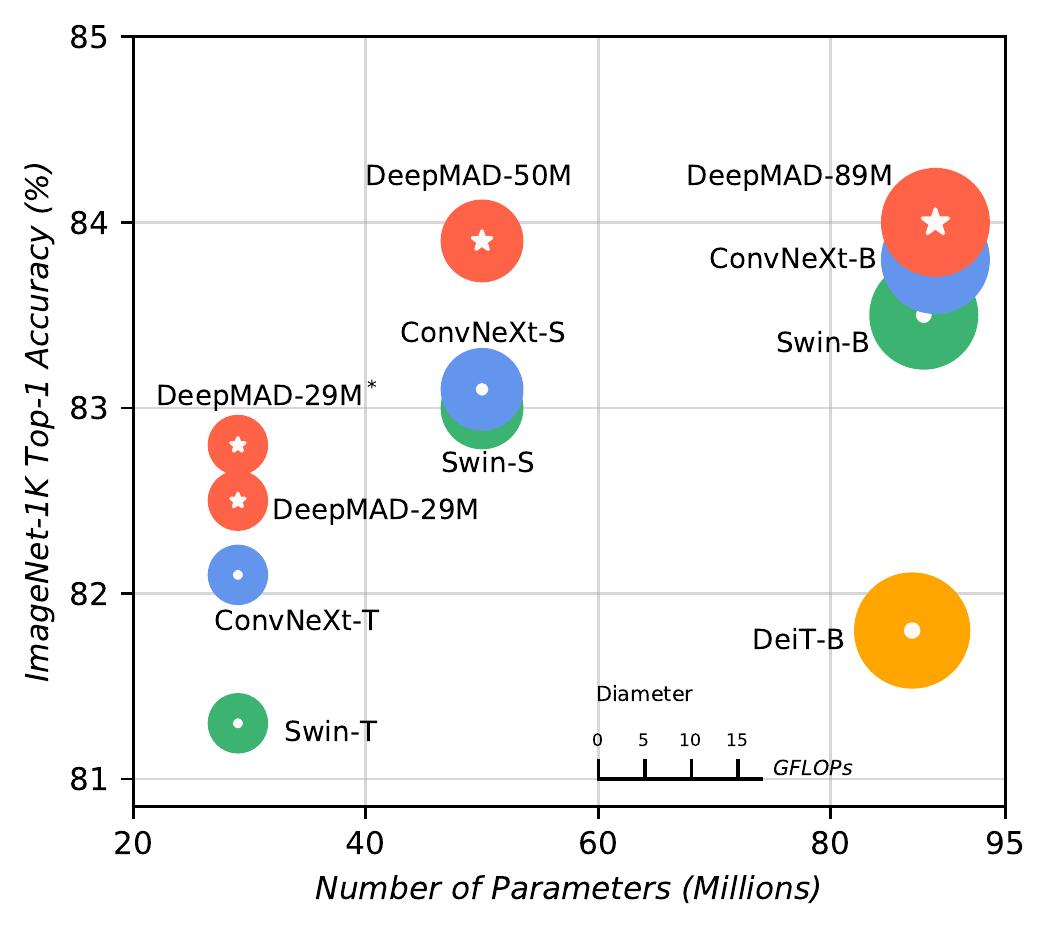}
  \caption{
  Comparison between DeepMAD models, Swin~\cite{liu2021swin} and ConvNeXt~\cite{liu2022convnet} on ImageNet-1k. DeepMAD achieves better performance than Swin and ConvNeXt with the same scales.
  }
  \label{fig:first_blood}
\end{figure}

Convolutional neural networks~(CNNs) have been the predominant computer vision models in the past decades~\cite{ka2012deepcnn, howard2017mobilenets,sandler2018mobilenetv2,efficientnet,liu2022convnet}. Until recently, the emergence of Vision Transformers (ViTs)~\cite{dosovitskiy2020vit, deit2021, liu2021swin} establishes a novel deep learning paradigm surpassing CNN models~\cite{deit2021, liu2021swin} thanks to the innovation of self-attention~\cite{attention2017} mechanism and other dedicated components~\cite{layernorm, shen2022lotteryvit, kong2021spvit, kong2022peeling, dong2022heatvit} in ViTs.

Despite the great success of ViT models in the 2020s, CNN models still enjoy many merits. First, CNN models do not require self-attention modules which require quadratic computational complexity in token size~\cite{mehta2021mobilevit}. Second, CNN models usually generalize better than ViT models when trained on small datasets~\cite{liu2022convnet}. In addition, convolutional operators have been well-optimized and tightly integrated on various hardware platforms in the industry, like IoT~\cite{han2020ofa}.


Considering the aforementioned advantages, recent researches try to revive CNN models using novel architecture designs \cite{yu2022metaformer,heo2021pit,liu2022convnet,ding2022scaling}. Most of these works adopt ViT components into CNN models, such as replacing the attention matrix with a convolutional counterpart while keeping the macrostructure of ViTs. After modifications, these modern CNN backbones are considerably different from the conventional ResNet-like CNN models. Although these efforts abridge the gap between CNNs and ViTs, designing such high-performance CNN models requires dedicated efforts in structure tuning and non-trivial prior knowledge of network design, therefore is time-consuming and difficult to generalize and customize.


In this work, a novel design paradigm named \textit{Mathematical Architecture Design}~(\textbf{DeepMAD}) is proposed, which designs high-performance CNN models \textbf{in a principled way}. DeepMAD is built upon the recent advances of deep learning theories~\cite{principlesofdeeplearning-2022, chan2021redunet, regnet2020}. To optimize the architecture of CNN models, DeepMAD innovates a constrained mathematical programming (MP) problem whose solution reveals the optimized structural parameters, such as the widths and depths of the network. Particularly, DeepMAD maximizes the differential entropy~\cite{zhenhong2022maedet, sk1997entropy, yy2020entropy, etj1957entropy,sunentropy,wang2023maximizing} of the network with constraints from the perspective of \textit{effectiveness}~\cite{principlesofdeeplearning-2022}.
The \textit{effectiveness} controls the information flow in the network which should be carefully tuned so that the generated networks are well behaved.
The dimension of the proposed MP problem in DeepMAD is less than a few dozen.
Therefore, it can be solved by off-the-shelf MP solvers nearly instantly on CPU. NO GPU is required and no deep model is created in memory\footnote{Of course, after solving the MP, training the generated DeepMAD models needs GPU}. This makes DeepMAD lightning fast even on CPU-only servers with a small memory footprint.
After solving the MP problem, the optimized CNN architecture is derived from the MP solution.

DeepMAD is a mathematical framework to design optimized CNN networks with strong theoretical guarantees and state-of-the-art (SOTA) performance. To demonstrate the power of DeepMAD, we use DeepMAD to optimize CNN architectures only using the conventional convolutional layers~\cite{si2015batchnorm,agarap2018deep} as building blocks. DeepMAD achieves comparable or better performance than ViT models of the same model sizes and FLOPs. Notably, DeepMAD achieves 82.8\% top-1 accuracy on ImageNet-1k with 4.5G FLOPs and 29M Params, outperforming ConvNeXt-Tiny (82.1\%)~\cite{liu2022convnet} and Swin-Tiny (81.3\%)~\cite{liu2021swin} at the same scale; DeepMAD also achieves 77.7\% top-1 accuracy at the same scale as ResNet-18~\cite{He2015resnet} on ImageNet-1k, which is 8.9\% better than He's original ResNet-18 (70.9\%) and is even comparable to He's ResNet-50 (77.4\%). The contributions of this work are summarized as follows:
\begin{compactitem}
    \item A Mathematical Architecture Design paradigm, DeepMAD, is proposed for high-performance CNN architecture design. 
    \item DeepMAD is backed up by modern deep learning theories~\cite{principlesofdeeplearning-2022, chan2021redunet, regnet2020}. It solves a constrained mathematical programming (MP) problem  to generate optimized CNN architectures. The MP problem can be solved on CPUs with a small memory footprint.
    \item DeepMAD achieves SOTA performances on multiple large-scale vision datasets, proving its superiority. Even only using the conventional convolutional layers, DeepMAD designs high-performance CNN models comparable to or better than ViT models of the same model sizes and FLOPs.
    \item DeepMAD is transferable across multiple vision tasks, including image classification, object detection, semantic segmentation and action recognition, with consistent performance improvements.
\end{compactitem}

\section{Related Works}

In this section, we briefly survey the recent works of modernizing CNN networks, especially the works inspired by transformer architectures. Then we discuss related works in information theory and theoretical deep learning.

\subsection{Modern Convolutional Neural Networks}

Convolutional deep neural networks are popular due to their conceptual simplicity and good performance in computer vision tasks. In most studies, CNNs are usually manually designed~\cite{He2015resnet, liu2022convnet, simonyan2014vgg, ding2022scaling, efficientnet, howard2017mobilenets}. These pre-defined architectures heavily rely on  human prior knowledge and are difficult to customize, for example, \xuan{tailored} to some given FLOPs/Params budgets. Recently, some works use AutoML~\cite{yao2022nasi, ming_zennas_iccv2021, chen2020tenas, sirui2019snas, chenxi2018pnas, mingxing2019mnasnet, changlin2020bsnas} to automatically generate high-performance CNN architectures. Most of these methods are data-dependent and require lots of computational resources. Even if one does not care about the computational cost of AutoML, the patterns generated by AutoML algorithms are difficult to interpret. It is hard to justify why such architectures are preferred and what theoretical insight we can learn from these results. Therefore, it is important to explore the architecture design in a principled way with clear theoretical motivation and human readability.

The Vision Transformer (ViT) is a rapid-trending topic in computer vision~\cite{dosovitskiy2020vit, deit2021, liu2021swin}. The Swin Transformer~\cite{liu2021swin} improves the computational efficiency of ViTs using a CNN-like stage-wise design. Inspired by Swin Transformer, recent researches combine CNNs and ViTs, leading to more efficient architectures~\cite{heo2021pit, liu2022convnet, ding2022scaling, liu2021swin, yu2022metaformer}. For example, MetaFormer~\cite{yu2022metaformer} shows that the attention matrix in ViTs can be replaced by a pooling layer. 
ConvNext~\cite{liu2022convnet} mimics the attention layer using depth-wise convolution and uses the same macro backbone as Swin Transformer~\cite{liu2021swin}.
RepLKNet~\cite{ding2022scaling} scales up the kernel sizes beyond 31$\times$31 to capture global receptive fields as attention.
All these efforts demonstrate that CNN models can achieve as good performance as ViT models when tuned carefully.
However, these modern CNNs require non-trivial prior knowledge when designing therefore are difficult to generalize and customize.

\subsection{Information Theory in Deep Learning}

Information theory is a powerful instrument for studying complex systems such as deep neural networks. The Principle of Maximum Entropy~\cite{sk1997entropy, etj1957entropy} is one of the most widely used principles in information theory. 
Several previous works~\cite{principlesofdeeplearning-2022,zhenhong2022maedet, am2018infodl, chan2021redunet, yy2020entropy} attempt to establish the connection between the information entropy and the neural network architectures.
For example, \cite{chan2021redunet} tries to interpret the learning ability of deep neural networks using subspace entropy reduction. \cite{am2018infodl} studies the information bottleneck in deep architectures and explores the entropy distribution and information flow in deep neural networks.
\cite{yy2020entropy} proposes the principle of maximal coding rate reduction for optimization.
\cite{zhenhong2022maedet} designs efficient object detection networks via maximizing multi-scale feature map entropy.
The monograph~\cite{principlesofdeeplearning-2022} analyzes the mutual information between different neurons in an MLP model.
In DeepMAD, the entropy of the model itself is considered  instead of the coding rate reduction as in~\cite{chan2021redunet}. The \textit{effectiveness} is also proposed to show that only maximizing entropy as in~\cite{zhenhong2022maedet} is not enough.



\section{Mathematical Architecture Design for MLP}   \label{sec:mlpdesign}

In this section, we study the architecture design for Multiple Layer Perceptron (MLP) using a novel mathematical programming (MP) framework. We then generalize this technique to CNN models in the next section.
To derive the MP problem for MLP, we first define the entropy of the MLP which controls its \textit{expressiveness}, followed by a constraint which controls its \textit{effectiveness}.
Finally, we maximize the entropy
 {objective function subject to the effectiveness constraint.}

\subsection{Entropy of MLP models}
\label{section:Entropy of MLP models}

Suppose that in an $L$-layer MLP $f(\cdot)$, the $i$-th layer has $w_i$ input channels and $w_{i+1}$ output channels. The output $\mathbf{x}_{i+1}$ and the input $\mathbf{x}_{i}$ are connected by $\mathbf{x}_{i+1} = \mathbf{M}_i \mathbf{x}_{i}$ where $\mathbf{M}_i \in \mathbb{R}^{w_{i+1} \times w_{i}}$ is trainable weights. Following the entropy analysis in~\cite{chan2021redunet}, the entropy of the MLP model $f(\cdot)$ is given in Theorem~\ref{thm:entroy-of-mlp}.

\begin{theorem} \label{thm:entroy-of-mlp}
The normalized Gaussian entropy upper bound of the MLP $f(\cdot)$ is 
\begin{align}
H_f= w_{L+1}
\sum_{i=1}^{L} \log(w_i) \label{eq:entropy-of-mlp}.
\end{align}
\end{theorem}
The proof is given in Appendix~\ref{sec:proof_entropy_of_mlp}.
The entropy measures the \textit{expressiveness} of the deep network~\cite{chan2021redunet,zhenhong2022maedet}. Following \textit{the Principle of Maximum Entropy}~\cite{sk1997entropy, etj1957entropy}, we propose to maximize the entropy of MLP under given computational budgets.

However, simply maximizing entropy defined in Eq.~(\ref{eq:entropy-of-mlp})  leads to an over-deep network because the entropy grows exponentially faster in depth than in width according to the Theorem~\ref{thm:entroy-of-mlp}.
An over-deep network is difficult to train and hinders effective information propagation~\cite{principlesofdeeplearning-2022}. This observation inspires us to look for another dimension in deep architecture design. This dimension is termed \textit{effectiveness} presented in the next subsection.

\subsection{Effectiveness Defined in MLP}    \label{section:mlp_effective}

An over-deep network can be considered as a chaos system that hinders effective information propagation. For a chaos system, when the weights of the network are randomly initialized, a small perturbation in low-level layers of the network will lead to an exponentially large perturbation in the high-level output of the network. During the back-propagation, the gradient flow cannot effectively propagate through the whole network. Therefore, the network becomes hard to train when it is too deep.

Inspired by the above observation, in DeepMAD we propose to control the depth of network. Intuitively, a 100-layer network is relatively too deep if its width is only 10 channels per layer or is relatively too shallow if its width is 10000 channels per layer. To capture this relative-depth intuition rigorously,  we import the metric termed network \textit{effectiveness} for MLP from the work~\cite{principlesofdeeplearning-2022}. Suppose that an MLP has $L$-layers and each layer has the same width $w$, the \textit{effectiveness} of this MLP is defined by
\begin{align}
\rho = L / w \label{eq:rho-L-W} \ .
\end{align}
Usually, $\rho$ should be a small constant. When $\rho \rightarrow 0 $, the MLP behaves like a single-layer linear model; when $\rho \rightarrow \infty$, the MLP is a chaos system. There is an optimal $\rho^*$ for MLP such that the mutual information between the input and the output are maximized~\cite{principlesofdeeplearning-2022}.

In DeepMAD, we propose to constrain the \textit{effectiveness} when designing the network. An unaddressed issue is that Eq.~(\ref{eq:rho-L-W}) assumes the MLP has uniform width but in practice, the width $w_i$ of each layer can be different. To address this issue, we propose to use the average width of MLP in Eq.~(\ref{eq:rho-L-W}).

\begin{proposition} \label{pro:average-width-of-mlp}
The average width of an $L$ layer MLP $f(\cdot)$ is defined by
\begin{align}
\bar{w}=(\prod_{i=1}^{L} w_i)^{1/L} = \exp \left( \frac{1}{L} \sum_{i=1}^{L} \log w_i \right) \ .  \label{eq_compute_width}
\end{align}
\end{proposition}

Proposition~\ref{pro:average-width-of-mlp} uses geometric average instead of arithmetic average of $w_i$ to define the average width of MLP. This definition is derived from the entropy definition in Eq.~(\ref{eq:entropy-of-mlp}). Please check Appendix~\ref{sec:proof_compute_width} for details.
In addition, geometric average is more reasonable than arithmetic average. Suppose an MLP has a zero width in some layer. Then the information cannot propagate through the network. Therefore, its ``equivalent width" should be zero. 

In real-world applications, the optimal value of $\rho$ depends on the building blocks. We find that $\rho \in [0.1, 2.0]$ usually gives good results in most vision tasks.


\section{Mathematical Architecture Design for CNN}   \label{sec:cnndesign}

In this section, the definitions of entropy and the {\it effectiveness} are generalized from MLP to CNN. Then three empirical guidelines are introduced inspired by the best engineering practice. At last, the final mathematical formulation of DeepMAD is presented.

\subsection{From MLP to CNN}
\label{section: From MLP to CNN}

A CNN operator is essentially a matrix multiplication with a sliding window. Suppose that in the $i$-th CNN layer, the number of input channels is $c_i$, the number of output channels is $c_{i+1}$, the kernel size is $k_i$, group is $g_i$. Then this CNN operator is equivalent to a matrix multiplication $W_i \in \mathbb{R}^{c_{i+1}\times c_i k_i^2 /g_i}$. Therefore, the ``width" of this CNN layer is projected to $c_i k_i^2 / g_i$ in Eq.~(\ref{eq:entropy-of-mlp}).

A new dimension in CNN feature maps is the resolution $r_i \times r_i$ at the $i$-th layer. To capture this, we propose the following definition of entropy for CNN networks.
\begin{proposition}
For an $L$-layer  CNN network $f(\cdot)$ parameterized by $\{c_i,k_i,g_i,r_i\}_{i=1}^{L}$, its entropy is defined by
\begin{align}
    H_L \triangleq \log(r_{L+1}^2 c_{L+1}) \sum_{i=1}^{L} \log(c_i k_i^2 / g_i) \ . \label{eq:entropy-for-cnn}
\end{align}
\end{proposition}

In Eq.~(\ref{eq:entropy-for-cnn}), we use a similar definition of entropy as in Eq.~(\ref{eq:entropy-of-mlp}). We use $\log(r_{L+1}^2 c_{L+1})$ instead of $(r_{L+1}^2 c_{L+1})$ in Eq.~(\ref{eq:entropy-for-cnn}). This is because a nature image is highly compressible so the entropy of an image or feature map does not scale up linearly in its volume $O(r_i^2 \times c_i)$. Inspired by~\cite{The-statistics-of-natural-images}, taking logarithms can better formulate the ground-truth entropy for natural images.

\subsection{Three Empirical Guidelines} \label{sec:three-additional-rules}

We find that the following three heuristic rules are beneficial to architecture design in DeepMAD. These rules are inspired by the best engineering practices.

\begin{itemize}
    \item \textbf{Guideline 1. Weighted Multiple-Scale Entropy}\quad CNN networks usually contain down-sampling layers which split the network into several stages. Each stage captures features at a certain scale. To capture the entropy at different scales, we use a weighted summation to ensemble entropy of the last layer in each stage to obtain the entropy of the network as in~\cite{zhenhong2022maedet}.
    \item  \textbf{Guideline 2. Uniform Stage Depth}\quad We require the depth of each stage to be uniformly distributed as much as possible. We use the variance of depths to measure the uniformity of depth distribution.
    \item \textbf{Guideline 3. Non-Decreasing Number of Channels} \quad We require that channel number of each stage is non-decreasing along the network depth. This can prevent high-level stages from having small widths. This guideline is also a common practise in a lot of manually designed networks.
\end{itemize}




\subsection{Final DeepMAD Formula}
\label{section:the proposed formula}
We gather everything together and present the final mathematical programming problem for DeepMAD. Suppose that we aim to design an $L$-layer CNN model $f(\cdot)$ with $M$ stages. The entropy of the $i$-th stage is denoted as $H_i$ defined in Eq.~(\ref{eq:entropy-for-cnn}). Within each stage, all blocks use the same structural parameters (width, kernel size, etc.). The width of each CNN layer is defined by $w_i=c_i k_i^2 / g_i$. The depth of each stage is denoted as $L_i$ for $i=1,2, \cdots, M$.  We propose to optimize $\{w_i, L_i\}$ via the following mathematical programming (MP) problem:

\begin{equation}
    \label{eq_mp_problem}
    \begin{aligned}
        \max_{w_i, L_i} \quad & \sum_{i=1}^{M} \alpha_i H_i - \beta Q, \\
        \mathrm{s.t.} \quad   & L \cdot (\prod_{i=1}^{L} w_i)^{-1/L}  \leq \rho_0, \\
                              & \mathrm{FLOPs}[f(\cdot)] \leq \mathrm{budget}, \\
                              & \mathrm{Params}[f(\cdot)] \leq \mathrm{budget}, \\
                              & Q \triangleq \exp[\mathrm{Var}(L_1,L_2, \cdots, L_M)], \\
                              & w_1 \leq w_2 \leq \cdots \leq w_L. \\
    \end{aligned}
\end{equation}

In the above MP formulation, $\{\alpha_i, \beta, \rho_0\}$ are hyper-parameters. $\{\alpha_i\}$ are the weights of entropies at different scales. For CNN models with 5 down-sampling layers, $\{\alpha_i\}=\{1,1,1,1,8\}$ is suggested in most vision tasks. $Q$ penalizes the objective function if the network has non-uniform depth distribution across stages. We set $\beta=10$ in our experiments. $\rho_0$ controls the \textit{effectiveness} of the network whose value is usually tuned in range $[0.1, 2.0]$. The last two inequalities control the computational budgets. This MP problem can be easily solved by off-the-shelf solvers for constrained non-linear programming~\cite{matlab, ea1996}.

\section{Experiments}

Experiments are developed at three levels.
First, the relationship between the model accuracy and the model \textit{effectiveness} is investigated on CIFAR-100~\cite{cifar100} to verify our effective theory in Section~\ref{section:the proposed formula}. Then, DeepMAD is used to design better ResNets and mobile networks. To demonstrate the power of DeepMAD, we design SOTA CNN models using DeepMAD with the conventional convolutional layers. Performances on ImageNet-1K~\cite{imagenet} are reported with comparison to popular modern CNN and ViT models.

Finally, the CNN models designed by DeepMAD are transferred to multiple down-streaming tasks, such as MS COCO~\cite{cocodataset} for object detection, ADE20K~\cite{ade20k} for semantic segmentation and UCF101~\cite{soomro2012ucf101} / Kinetics400~\cite{kinetics400} for action recognition. Consistent performance improvements demonstrate the excellent transferability of DeepMAD models.


\subsection{Training Settings}
Following previous works~\cite{Bag_of_tricks_Xie2019,resnet_strikes_back},
SGD optimizer with momentum 0.9 is adopted to train DeepMAD models.
The weight decay is 5e-4 for CIFAR-100 dataset and 4e-5 for ImageNet-1k. The initial learning rate is 0.1 with batch size of 256. We use cosine learning rate decay~\cite{loshchilov2016sgdr} with 5 epochs of warm-up. The number of training epochs is 1,440 for CIFAR-100 and 480 for ImageNet-1k. All experiments use the following data augmentations~\cite{pham2018efficient}: mix-up~\cite{zhang2020overcoming}, label-smoothing~\cite{szegedy2016rethinking}, random erasing~\cite{zhong2020random}, random crop/resize/flip/lighting, and Auto-Augment~\cite{cubuk2018autoaugment}.

\subsection{Building Blocks}
To align with ResNet family~\cite{He2015resnet},  Section~\ref{sec:DeepMAD-for-resent} uses the same building blocks as ResNet-50.
To align with ViT models~\cite{dosovitskiy2020vit, deit2021, liu2021swin}, DeepMAD  uses MobileNet-V2~\cite{howard2017mobilenets} blocks followed by SE-block~\cite{hu2018squeeze} as in EfficientNet~\cite{efficientnet} to design high performance networks.

\subsection{Effectiveness on CIFAR-100}
\label{section:Validations on CIFAR-100}
The {\it effectiveness} $\rho$ is an important hyper-parameter in DeepMAD. This experiment demonstrate how $\rho$ affects the architectures in DeepMAD. To this end, 65 models are randomly generated using ResNet blocks, with different depths and widths. All models have the same FLOPs (0.04G) and Params (0.27M) as ResNet-20~\cite{resnet20cifar} for CIFAR-100. The \textit{effectiveness} $\rho$ varies in range $[0.1, 1.0]$.

\begin{figure}[t]
  \centering
  \includegraphics[width=1.0\linewidth]{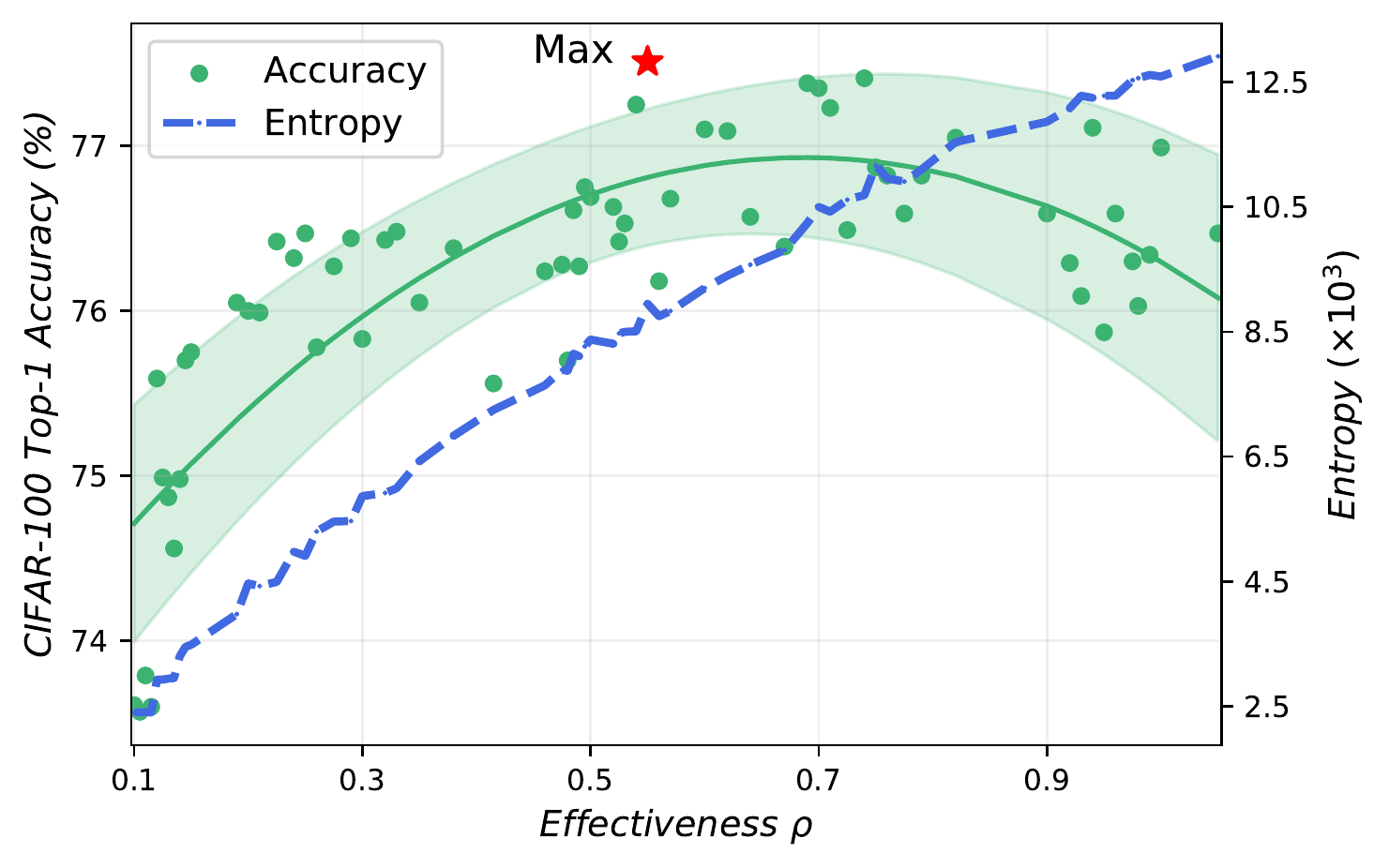}
  \caption{
  {\it Effectiveness} $\rho$ v.s. top-1 accuracy and entropy of each generated model on CIFAR-100. The best model is marked by a star. The entropy increases with $\rho$ monotonically but the model accuracy does not. The optimal $\rho^*\approx 0.5$. 
  }
  \label{fig:rho_ablation_cifar_1}
\end{figure}

These randomly generated models are trained on CIFAR-100.
The {\it effectiveness} $\rho$, top-1 accuracy and network entropy for each model are plotted in Figure~\ref{fig:rho_ablation_cifar_1}.
We can find that the entropy increases with $\rho$ monotonically.
This is because the larger the $\rho$ is, the deeper the network is, and thus the greater the entropy as described in Section~\ref{section:Entropy of MLP models}.
However, as shown in Figure~\ref{fig:rho_ablation_cifar_1}, the model accuracy does not always increase  with $\rho$ and entropy. When $\rho$ is small, the model accuracy is proportional to the model entropy; when $\rho$ is too large, such relationship no longer exists. Therefore, $\rho$ should be contrained in a certain ``effective'' range in DeepMAD.

\begin{figure}[t]
  \centering
  \includegraphics[width=0.9\linewidth]{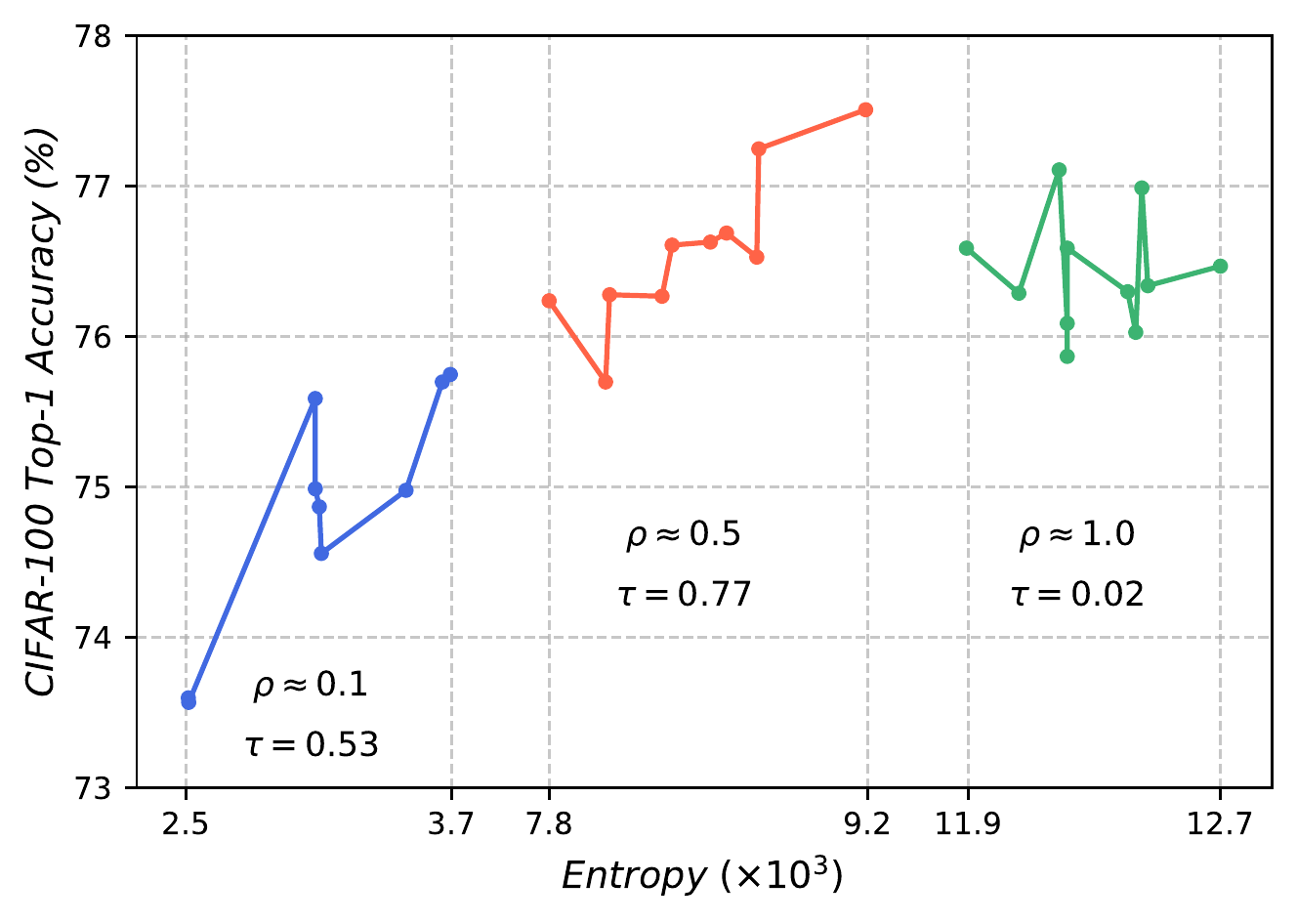}
    \caption{
    The architectures around $\rho=\{0.1, 0.5, 1.0\}$ are selected and grouped by $\rho$. Kendall coefficient $\tau$~\cite{abdi2007kendall} is used to measure the correlation.
    }
  \label{fig:rho_ablation_cifar_2}
\end{figure}

Figure~\ref{fig:rho_ablation_cifar_2} gives more insights into the effectiveness hypothesis in Section~\ref{section:mlp_effective}.
The architectures around $\rho=\{0.1, 0.5, 1.0\}$ are selected and grouped by $\rho$. When $\rho$ is small ($\rho$ is around 0.1), the network is effective in information propagation so we observe a strong correlation between network entropy and network accuracy.
But these models are too shallow to obtain high performance.
When $\rho$ is too large ($\rho\approx 1.0$), the network approaches a chaos system therefore no clear correlation between network entropy and network accuracy.
When $\rho$ is around 0.5, the network can achieve the best performance and the correlation between the network entropy and network accuracy reaches 0.77.

\subsection{DeepMAD for ResNet Family} \label{sec:DeepMAD-for-resent}

ResNet family is one of the most popular and classic CNN models in deep learning. We use DeepMAD to re-design ResNet and show that the DeepMAD can generate much better ResNet models.
The \textit{effectiveness} $\rho$ for those original ResNets is computed for easy comparison.
First, we use DeepMAD to design a new architecture DeepMAD-R18 which has the same model size and FLOPs as ResNet-18. $\rho$ is tuned in range \{0.1, 0.3, 0.5, 0.7\} for DeepMAD-R18. $\rho=0.3$ gives the best architecture.
Then, $\rho=0.3$ is fixed in the design of DeepMAD-R34 and DeepMAD-R50 which align with ResNet-34 and Resnet-50 respectively.
As shown in Table~\ref{tab:res}, compared to He's original results, DeepMAD-R18 achieves 6.8\% higher accuracy than ResNet-18 and is even comparable to ResNet-50. Besides, DeepMAD-R50 achieves 3.2\% better accuracy than the ResNet-50.
To ensure the fairness in comparison, the performances of ResNet family under the fair training setting are reported.
With our training recipes, the accuracies of ResNet models improved around $1.5\%$.
DeepMAD still outperforms the ResNet family by a large margin when both are trained fairly.
The inferior performance of ResNet family can be explained by their small $\rho$ which limits their model entropy. This phenomenon again validates our theory discussed in Section~\ref{section:the proposed formula}.

\begin{table}[t]
  \centering
  \resizebox{1.0\linewidth}{!}{
  \begin{tabular}{lcccc}
    \toprule
    Model  & \# Param. & FLOPs & $\rho$ &  Acc. (\%) \\
    \midrule
    ResNet-18~\cite{He2015resnet} & 11.7\,M & 1.8\,G & 0.01  & 70.9 \\
    ResNet-18$\dagger$ & 11.7\,M & 1.8\,G & 0.01  & 72.2 \\
    DeepMAD-R18 & 11.7\,M & 1.8\,G & 0.1 & 76.9 \\
    \rowcolor{light-gray} DeepMAD-R18 & 11.7\,M & 1.8\,G & \textbf{0.3} & \textbf{77.7} \\
    DeepMAD-R18 & 11.7\,M & 1.8\,G & 0.5 & 77.5 \\
    DeepMAD-R18 & 11.7\,M & 1.8\,G & 0.7 & 75.7 \\
    \midrule
    ResNet-34~\cite{He2015resnet} & 21.8\,M & 3.6\,G & 0.02 & 74.4 \\
    ResNet-34$\dagger$ & 21.8\,M & 3.6\,G & 0.02 & 75.6 \\
    \rowcolor{light-gray} DeepMAD-R34 & 21.8\,M & 3.6\,G & 0.3 & \textbf{79.7} \\
    \midrule
    ResNet-50~\cite{He2015resnet} & 25.6\,M & 4.1\,G & 0.09 & 77.4 \\
    ResNet-50$\dagger$ & 25.6\,M & 4.1\,G & 0.09 & 79.3 \\
    \rowcolor{light-gray} DeepMAD-R50 & 25.6\,M & 4.1\,G & 0.3 & \textbf{80.6} \\
    \bottomrule
  \end{tabular}}
  \caption{
  DeepMAD v.s. ResNet on ImageNet-1K, using ResNet building block. 
  $\dagger$: model trained by our pipeline.
  $\rho$ is tuned for DeepMAD-R18.
  DeepMAD achieves consistent improvements compared with ResNet18/34/50 with the same Params and FLOPs.
  }
  \label{tab:res}
\end{table}

\subsection{DeepMAD for Mobile CNNs} \label{sec:DeepMAD-for-mobile-device}

We use DeepMAD to design mobile CNN models for further exploration.
Following previous works, MobileNet-V2 block with SE-block are used to build new models. $\rho$ is tuned at EfficientNet-B0 scale in the range of \{0.3, 0.5, 1.0, 1.5, 2.0\} for DeepMAD-B0, and $\rho=0.5$ achieves the best result. Then, we transfer the optimal $\rho$ for DeepMAD-B0 to DeepMAD-MB.
As shown in Table~\ref{tab:depthwise}, the DeepMAD-B0 achieves 76.1\% top-1 accuracy which is comparable with the EfficientNet-B0~(76.3\%).
It should be noted that EfficientNet-B0 is designed by brute-force grid search which takes around 3800 GPU days~\cite{FBNetV2}.
The performance of the DeepMAD-B0 is comparable to the EfficientNet-B0 by simply solving an MP problem \xuan{on CPU} in a few minutes.
Aligned with MobileNet-V2 on Params and FLOPs, DeepMAD-MB achieves 72.3\% top-1 accuracy which is 0.3\% higher in accuracy.

\begin{table}[t]
  \centering
  \resizebox{1.0\linewidth}{!}
  {
  \begin{tabular}{lcccc}
    \toprule
    Model & \# Param. & FLOPs & $\rho$ & Acc. (\%) \\
    \midrule
    EffNet-B0~\cite{efficientnet} & 5.3\,M & 390\,M & 0.6 & 76.3 \\
    DeepMAD-B0 & 5.3\,M & 390\,M & 0.3 & 74.3 \\
    \rowcolor{light-gray} DeepMAD-B0 & 5.3\,M & 390\,M & \textbf{0.5} & \textbf{76.1} \\
    DeepMAD-B0 & 5.3\,M & 390\,M & 1.0 & 75.9 \\
    DeepMAD-B0 & 5.3\,M & 390\,M & 1.5 & 75.7 \\
    DeepMAD-B0 & 5.3\,M & 390\,M & 2.0 & 74.9 \\
    \midrule
    MobileNet-V2~\cite{howard2017mobilenets} & 3.5\,M & 320\,M & 0.9 & 72.0 \\
    \rowcolor{light-gray} DeepMAD-MB & 3.5\,M & 320\,M & 0.5 & \textbf{72.3} \\
    \bottomrule
  \end{tabular}}
  \caption{DeepMAD under mobile setting. Top-1 accuracy on ImageNet-1K. $\rho$ is tuned for DeepMAD-B0.}
  \label{tab:depthwise}
\end{table}

\subsection{DeepMAD for SOTA}

\begin{table}[t]
  \centering
  \begin{tabular}{llllll}
    \toprule
    Model & \makecell{Res} & Params & FLOPs & Acc (\%) \\
    \midrule
    ResNet-50~\cite{He2015resnet} & 224 & 26\,M & 4.1\,G & 77.4 \\
    DeiT-S~\cite{deit2021} & 224 & 22\,M & 4.6\,G  & 79.8 \\
    PVT-Small~\cite{wang2021pyramid} & 224 & 25\,M & 3.8\,G  & 79.8 \\
    Swin-T~\cite{liu2021swin} & 224 & 29\,M & 4.5\,G  & 81.3 \\
    TNT-S~\cite{han2021transformer} & 224 & 24\,M & 5.2\,G  & 81.3 \\
    T2T-ViT{$_t$}-14~\cite{yuan2021tokens} & 224 & 22\,M & 6.1\,G  & 81.7 \\
    ConvNeXt-T~\cite{liu2022convnet} & 224 & 29\,M & 4.5\,G  & 82.1 \\
    SLaK-T~\cite{liu2022more} & 224 & 30\,M & 5.0\,G  & 82.5 \\
    \rowcolor{light-gray} DeepMAD-29M & 224 & 29\,M & 4.5\,G &  \textbf{82.5} \\
    \rowcolor{light-gray} DeepMAD-29M$^{*}$ & 288 & 29\,M & 4.5\,G &  \textbf{82.8} \\
    \midrule
    ResNet-101~\cite{He2015resnet} & 224 & 45\,M & 7.8\,G & 78.3 \\
    ResNet-152~\cite{He2015resnet} & 224 & 60\,M & 11.5\,G & 79.2 \\
    PVT-Large~\cite{wang2021pyramid} & 224 & 61\,M & 9.8\,G  & 81.7 \\
    T2T-ViT{$_t$}-19~\cite{yuan2021tokens} & 224 & 39\,M & 9.8\,G  & 82.2 \\
    T2T-ViT{$_t$}-24~\cite{yuan2021tokens} & 224 & 64\,M & 15.0\,G  & 82.6 \\
    TNT-B~\cite{han2021transformer} & 224 & 66\,M & 14.1\,G  & 82.9 \\
    Swin-S~\cite{liu2021swin} & 224 & 50\,M & 8.7\,G  & 83.0 \\
    ConvNeXt-S~\cite{liu2022convnet} & 224 & 50\,M & 8.7\,G  & 83.1 \\
    SLaK-S~\cite{liu2022more} & 224 & 55\,M & 9.8\,G  & 83.8 \\
    \rowcolor{light-gray} DeepMAD-50M & 224 & 50\,M & 8.7\,G  & \textbf{83.9} \\
    \midrule
    DeiT-B/16~\cite{deit2021} & 224 & 87\,M & 17.6\,G  & 81.8 \\
    RepLKNet-31B~\cite{ding2022scaling} & 224 & 79\,M & 15.3\,G  & 83.5 \\
    Swin-B~\cite{liu2021swin} & 224 & 88\,M & 15.4\,G  & 83.5 \\
    ConvNeXt-B~\cite{liu2022convnet} & 224 & 89\,M & 15.4\,G  & 83.8 \\
    SLaK-B~\cite{liu2022more} & 224 & 95\,M & 17.1\,G  & 84.0 \\
    \rowcolor{light-gray} DeepMAD-89M & 224 & 89\,M & 15.4\,G  & \textbf{84.0} \\
    \bottomrule
  \end{tabular}
  \caption{
  DeepMAD v.s. SOTA ViT and CNN models on ImageNet-1K. $\rho=0.5$ for all DeepMAD models.
  DeepMAD-29M$^{*}$: uses 288x288 resolution while the Params and FLOPs keeps the same as DeepMAD-29M.
  }
  \label{tab:sota}
\end{table}

We use DeepMAD to design a SOTA CNN model for ImageNet-1K classification. The conventional MobileNet-V2 building block with SE module is used.
This DeepMAD network is aligned with Swin-Tiny~\cite{liu2021swin} at 29M Params and 4.5G FLOPs therefore is labeled as DeepMAD-29M.
\cross{This work did not aim at bigger models considering computational resource constraints for training and the model practicality.}
As shown in Table~\ref{tab:sota},
DeepMAD-29M outperforms or is comparable to SOTA ViT models as well as recent modern CNN models.
DeepMAD-29M achieves 82.5\%, which is 2.7\% higher accuracy than DeiT-S~\cite{deit2021} and 1.2\% higher accuracy than the Swin-T~\cite{liu2021swin}.
Meanwhile, DeepMAD-29M is 0.4\% higher than the ConvNeXt-T~\cite{liu2022convnet} which is inspired by the transformer architecture.
DeepMAD also designs networks with larger resolution (288), DeepMAD-29M$^*$, while keeping the FLOPs and Params not changed.
DeepMAD-29M$^*$ reaches 82.8\% accuracy and is comparable to Swin-S~\cite{liu2021swin} and ConvNeXt-S~\cite{liu2022convnet} with nearly half of their FLOPs.
\xuan{Deep-MAD also achieves better performance on small and base level. Especially, DeepMAD-50M can achieve even better performance than ConvNeXt-B with nearly half of its scale.} 
It proves only with the conventional convolutional layers as building blocks, Deep-MAD achieves comparable or better performance than ViT models.

\subsection{Downstream Experiments}

To demonstrate the transferability of models designed by DeepMAD, the models solved by DeepMAD play as the backbones on downstream tasks including object detection, semantic segmentation and action recognition.

\paragraph{Object Detection on MS COCO}
MS COCO is a widely used dataset in object detection.
It has 143K images and 80 object categories.
The experiments are evaluated on MS COCO~\cite{cocodataset} with the official training/testing splits.
The results in Table~\ref{tab:detection} are evaluated on val-2017.
We use two detection frameworks, FCOS~\cite{wang2020fcos} and GFLV2~\cite{li2020gflv2}, implemented by mmdetection~\cite{mmdetection}.
The DeepMAD-R50 model plays as the backbone of these two detection frameworks.
The models are initialized with pre-trained weights on ImageNet and trained for 2X~(24 epoches).
The multi-scale training trick is also used for the best performance.
As shown in Table~\ref{tab:detection}, DeepMAD-R50 achieves 40.0 AP with FCOS~\cite{wang2020fcos} , which is 1.5 AP higher than ResNet-50.
It also achieves 44.9 AP with GFLV2~\cite{li2020gflv2}, which is 1.0 AP higher than ResNet-50 again.
The performance gain without introducing more Params and FLOPs proves the superiority of DeepMAD on network design.

\paragraph{Semantic segmentation on ADE20K}
ADE20K~\cite{ade20k} dataset is broadly used in semantic segmentation tasks.
It has 25k images and 150 semantic categories. 
The experiments are evaluated on ADE20K~\cite{ade20k} with the official training/testing splits.
The results in Table~\ref{tab:segmentation} are reported on testing part using mIoU.
UperNet~\cite{upernet} in mmseg~\cite{mmseg2020} is chosen as the segmentation framework.
As shown in the first block of Table~\ref{tab:segmentation}, DeepMAD-R50 achieves 45.6 mIoU on testing, which is 2.8 mIoU higher than ResNet-50, and even 0.8 mIoU higher than ResNet-101.
To compare to ViT and transformer-inspired models, DeepMAD-29M$^*$ is used as the backbone in UperNet.
As shown in the last block of Table~\ref{tab:segmentation}, DeepMAD-29M$^*$ achieves 46.9 mIoU on testing, which is 1.1 mIoU higher than Swin-T and 0.2 mIoU higher than ConvNeXt-T, with the same model size and less computation cost.
It proves the advantage of CNN models designed by DeepMAD compared to transformer-based or transformer-inspired models.

\begin{table}[t]
  \centering
  {
      \begin{tabular}{lcccc}
        \toprule
        Backbone  & \# Param. & FLOPs & AP \\
        \midrule
        \textbf{FCOS} \\
        ResNet-50 & 23.5\,M & 84.1\,G & 38.5 \\
        \rowcolor{light-gray} DeepMAD-R50 & 24.2\,M & 83.2\,G & \textbf{40.0} \\
        \midrule
        \textbf{GFLV2} \\
        ResNet-50  & 23.5\,M & 84.1\,G & 43.9 \\
        \rowcolor{light-gray} DeepMAD-R50 & 24.2\,M  & 83.2\,G & \textbf{44.9} \\
        \bottomrule
      \end{tabular}
  } 
  \caption{
  DeepMAD for object detection and instance segmentation on MS COCO~\cite{cocodataset} with  GFLV2~\cite{li2020gflv2}, FCOS~\cite{wang2020fcos}, Mask R-CNN~\cite{He2020MaskR} and Cascade Mask R-CNN~\cite{Cai2018CascadeRD} frameworks. Backbones are pre-trained on ImageNet-1K. FLOPs and Params are counted for Backbone.
}
  \label{tab:detection}

\end{table}

\begin{table}[t]
  \centering
  \begin{tabular}{lccc}
    \toprule
    Backbone  & \# Param. & FLOPs & mIoU \\
    \midrule
    ResNet-50  & 23.5\,M & 86.3\,G & 42.8 \\
    ResNet-101  & 42.5\,M & 164.3\,G & 44.8 \\
    \rowcolor{light-gray} DeepMAD-R50  & 24.2\,M & 85.2\,G & \textbf{45.6} \\
    \midrule
    Swin-T & 27.5\,M & 95.8\,G & 45.8 \\
    ConvNeXt-T  & 27.8\,M & 93.2\,G & 46.7 \\
    \rowcolor{light-gray} DeepMAD-29M$^{*}$ & 26.5\,M & 55.5\,G & \textbf{46.9} \\
    \bottomrule
  \end{tabular}
  \caption{
  DeepMAD for semantic segmentation on ADE20K~\cite{ade20k}. All models are pre-trained on the ImageNet-1K and then fine-tuned using UperNet~\cite{upernet} framework. FLOPs and Params are counted for Backbone.
  }
  \label{tab:segmentation}

\end{table}

\begin{table}[t]

  \centering
  \begin{tabular}{lccc}
    \toprule
    Backbone  & \# Param. & FLOPs & Acc. (\%) \\
    \midrule
    \textbf{UCF-101} & & & \\
    ResNet-50 & 23.5\,M & 7.3\,G & 83.0 \\
    \rowcolor{light-gray} DeepMAD-R50  & 24.2\,M & 7.3\,G & \textbf{86.9} \\
    \midrule
    \textbf{Kinetics-400} & & & \\
    ResNet-50 & 23.5\,M & 7.3\,G & 70.6 \\
    \rowcolor{light-gray} DeepMAD-R50  & 24.2\,M & 7.3\,G & \textbf{71.6} \\
    \bottomrule
  \end{tabular}
  \caption{
  DeepMAD for action recognition on UCF-101~\cite{soomro2012ucf101} and Kinetics-400~\cite{kinetics400} with the TSN~\cite{wang2016temporal} framework. Backbones are pre-trained on the ImageNet-1K. FLOPs and Params are counted for Backbone.
  }
  \label{tab:action}
\end{table}

\paragraph{Action recognition on UCF101 and Kinetics400}
The UCF101~\cite{soomro2012ucf101} dataset contains 13,320 video clips, covering 101 action classes.
The Kinetics400~\cite{kinetics400} dataset contains 400 human action classes, with more than 400 video clips for each action.
They are both widely used in action recognition tasks.
The results in Table~\ref{tab:action} are reported on the testing part using top-1 accuracy.
TSN~\cite{wang2016temporal} is adopted in mmaction~\cite{2020mmaction2} on UCF101 with split1 and Kinetics400 with official training/testing splits.
As shown in Table~\ref{tab:action}, DeepMAD-R50 achieves 86.9\% accuracy on UCF101 which is 3.9\% higher than ResNet-50, and achieves 71.6\% accuracy on Kinetics400 which is 1.0\% higher than ResNet-50, with the same model size and computation cost.
It shows that the models solved by DeepMAD can also be generalized to the recognition task on video datasets.




\subsection{Ablation Study}
In this section, we ablate important hyper-parameters and empirical guidelines in DeepMAD, with image classification on ImageNet-1K and object detection on COCO dataset.
The \textbf{complexity comparison} is in Appendix~\ref{sec:compare_to_nas_results}.

\paragraph{Ablation on entropy weights}
We generate networks with conventional convolution building blocks and different weight ratios ${\alpha_i}$.
The ratio $\alpha_5$ is tuned in \{1, 8 16\} while the others are set to 1 as in~\cite{zhenhong2022maedet}. As shown in Table~\ref{tab:stage_ratio_ablation}, larger final stage weight can improve the performance on image classification task, while a smaller one can improve the performance on downstream task (object detection).
For different tasks, additional improvements can be obtained by fine-tuning $\alpha_5$.
However, this work uses $\alpha_5=8$ setting as the balance between the image classification task and object detection task.
The experiments above have verified the advantage of DeepMAD on different tasks with global $\alpha_5$.


\paragraph{Ablation on the three empirical guidelines}
We generate networks using Mobilenet-V2 block with SE module and remove one of the three empirical guidelines discussed in Section~\ref{sec:three-additional-rules} at each time to explore their influence.
As shown in Table~\ref{tab:rules_ablation}, removing any one of the three guidelines will degrade the performance of the model. Particularly, the third guideline is the most critical one for DeepMAD.

\section{Limitations}




As no research is perfect, DeepMAD has several limitations as well.
First, three empirical guidelines discussed in Section~\ref{sec:three-additional-rules} do not have strong theoretical foundation. Hopefully, they can be removed or replaced in the future.
\cross{after we have more breakthroughs in deep learning theories.}
Second, DeepMAD has several hyper-parameters to tune, such as $\{\alpha_i\}$ and $\{\beta, \rho\}$.
\cross{These hyper-parameters are double-bladed since they can embed human prior knowledge in DeepMAD but can be tedious to tune if one does not have such knowledge.}
Third, DeepMAD focuses on conventional CNN layers at this stage while there are many more powerful and more modern building blocks such as transformers. It is potentially possible to generalize DeepMAD to these building blocks as well in future works.
\cross{since DeepMAD is built upon information theory and generic deep learning theory that are not tied to a specific network architecture.}

\begin{table}[t]
  \centering
  \begin{tabular}{lcccc}
    \toprule
    Model  & \# Param. & FLOPs & $\alpha_5$ &  Acc. (\%) \\
    \midrule
    DeepMAD-R18 & 11.7\,M & 1.8\,G & 1 & 76.7 \\
    DeepMAD-R18 & 11.7\,M & 1.8\,G & 8 & 77.7 \\
    DeepMAD-R18 & 11.7\,M & 1.8\,G & 16 & \textbf{78.7} \\
    \midrule
    Backbone  & \# Param. & FLOPs & $\alpha_5$ &  AP (\%) \\
    \midrule
    DeepMAD-R18 & 9.8\,M & 37.0\,G & 1 & \textbf{35.1} \\
    DeepMAD-R18 & 11.0\,M & 36.8\,G & 8 & 34.6 \\
    DeepMAD-R18 & 11.1\,M & 36.8\,G & 16 & 34.1 \\
    \bottomrule
  \end{tabular}
  \caption{
  The performance on ImageNet-1k and COCO dataset of DeepMAD-R18 with different final stage weight $\alpha_5$.
  $\alpha_5=8$ balances the good performance between classification and object detection and is adopted in this work.
  }
  \label{tab:stage_ratio_ablation}
\end{table}

\begin{table}[t]
  \centering
  \begin{tabular}{ccccc}
    \toprule
    
    Guideline1 & Guideline2 & Guideline3 & Acc. (\%) \\
    \midrule
    $\times$ & $\checkmark$ & $\checkmark$ & 73.5 \\
    $\checkmark$ & $\times$ & $\checkmark$ & 73.5 \\
    $\checkmark$ & $\checkmark$ & $\times$ & 73.1 \\
    $\checkmark$ & $\checkmark$ & $\checkmark$ & \textbf{73.7} \\
    \bottomrule
  \end{tabular}
  \caption{
  Top-1 accuracies on ImageNet-1K of DeepMAD-B0 with different combination of guidelines.
  The model designed with all three guidelines achieves the best results.
  All models are trained for 120 epochs.
  }
  \label{tab:rules_ablation}
\end{table}

\section{Conclusion}

We propose a pure mathematical framework DeepMAD for designing high-performance convolutional neural networks. The key idea of the DeepMAD is to maximize the network entropy while keeping network effectiveness bounded by a small constant. We show that DeepMAD can design SOTA CNN models that are comparable to or even better than ViT models and modern CNN models. To demonstrate the power of DeepMAD, we only use conventional convolutional building blocks, like ResNet block, and depth-wise convolution in MobileNet-V2. Without bells and whistles, DeepMAD achieves competitive performance using these old-school building blocks. This encouraging result implies that the full potential of the conventional CNN models has not been fully released due to the previous sub-optimal design.
Hope this work can attract more research attention to theoretical deep learning in the future.

\section{Acknowledgment}
This work was supported by Alibaba Research Intern Program and National Science Foundation CCF-1919117.

{\small
\bibliographystyle{ieee_fullname}
\bibliography{sections/reference}
}

\appendix
\clearpage
\newpage


\renewcommand{\floatpagefraction}{.99}
\begin{figure*}[t]
  \centering
  \includegraphics[width=1.0\linewidth]{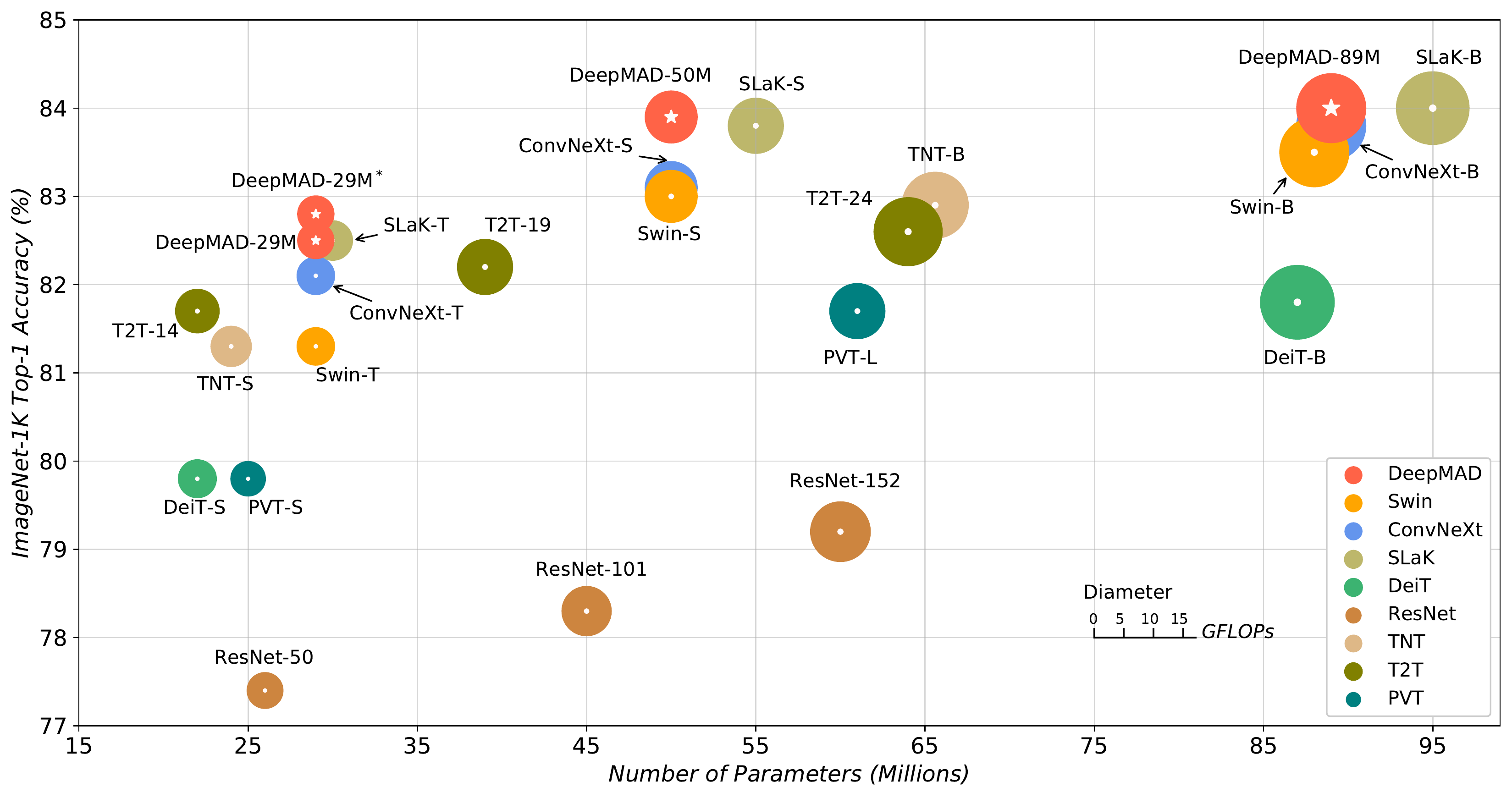}
    \caption{
     DeepMAD v.s. SOTA ViT and CNN models on ImageNet-1K. $\rho=0.5$ for all DeepMAD models.
     All DeepMAD models except DeepMAD-29M$^*$ is trained with 224 resolution.
     x-axis is the Params, the smaller the better.
     y-axis is the accuracy, the larger the better.
     }
  \label{fig:full_sota}
\end{figure*}


\section{Proof of Theorem~\ref{eq:entropy-of-mlp}} \label{sec:proof_entropy_of_mlp}







According to~\cite{cover1999elements,van2001reference,mood1950introduction,loeve2017probability}, we have the following  lemmas.
\begin{lemma}
The differential entropy $H(x)$ of random variable $x \sim{\mathcal{N}(0, \sigma)}$ is 
    \begin{equation}
        \label{eq:entropy}
        \begin{aligned}
            H(x) \propto log(\sigma^2).
        \end{aligned}
    \end{equation}
\end{lemma}

\begin{lemma} \label{lem:gaus-ent-upper-bound}
For any random variable $x$, its differential entropy $H(x)$ is  bounded by its Gaussian entropy upper bound
\begin{align}
    H(x) \leq c \log[\sigma^2(x)],
\end{align}
where $c>0$ is a universal constant, $\sigma(x)$ is the standard deviation of $x$.
\end{lemma}

\begin{lemma}
For N random variables $\{x_1, \cdots x_i, \cdots x_N\}$, the laws of expectation and variance of sum of random variables are
        \begin{align}
            \mathbb{E}(\sum_1^N x_i) &= \sum_1^N \mathbb{E}(x_i), \label{eq:Esum} \\
            \sigma^2(\sum_1^N x_i) &= \sum_1^N \sigma^2(x_i).  \label{eq:sigmasum}
        \end{align}
    The laws of expectation and variance of product of random variables are
    \begin{equation}
        \begin{aligned}
            \mathbb{E}(\prod_1^N x_i) &= \prod_1^N \mathbb{E}(x_i),  \label{eq:Eprod}\\
        \end{aligned}
    \end{equation}
    and 
    \begin{equation}
        \begin{split}
            \sigma^2(x_i x_j) = \sigma^2(x_i) \sigma^2(x_j) + \sigma^2(x_i)\mathbb{E}^2(x_j) + \\  \sigma^2(x_j)\mathbb{E}^2(x_i). \label{eq:sigmaprod}
        \end{split}
    \end{equation}
\end{lemma}

Suppose that in an $L$-layer MLP $f(\cdot)$, the $i$-th layer has $w_i$ input channels and $w_{i+1}$ output channels.
The trainable weights in $i$-th layer is denoted by $\mathbf{M}_i \in \mathbb{R}^{w_{i+1} \times w_{i}}$.
For simplicity, we assume that each element $\mathbf{x}_1^j$ in $\mathbf{x}_1$ and each element $\mathbf{M}_i^{j,k}$ in $\mathbf{M}_i$ follow the standard normal distribution, i.e.
\begin{align}
    \mathbf{x}_1^j & \sim{\mathcal{N}(0, 1)},     \label{eq:x_distribution} \\
    \mathbf{M}_i^{j,k} &\sim{\mathcal{N}(0, 1)}. \label{eq:w_distribution}
\end{align}

According to~\cref{eq:Esum,eq:sigmasum,eq:Eprod,eq:sigmaprod,eq:x_distribution}, in $i$-th layer, the output $\mathbf{x}_{i+1}$ and the input $\mathbf{x}_{i}$ are connected by $\mathbf{x}_{i+1} = \mathbf{M}_i \mathbf{x}_{i}$.
The expectation of $j$-th element in $\mathbf{x}_{i+1}$ is
\begin{equation}
    \label{eq:Ex}
    \begin{aligned}
        \mathbb{E}(\mathbf{x}_{i+1}^j) 
        &= \mathbb{E}(\sum_{k=1}^{w_i} \mathbf{M}_i^{jk} \mathbf{x}_i^{k}) \\
        &= \sum_{k=1}^{w_i} \mathbb{E}(\mathbf{M}_i^{jk} \mathbf{x}_i^{k}) \\
        &= \sum_{k=1}^{w_i} \mathbb{E}(\mathbf{M}_i^{jk}) \mathbb{E}(\mathbf{x}_i^{k}) \\
        &= 0.
    \end{aligned}
\end{equation}

According to ~\cref{eq:Esum,eq:sigmasum,eq:Eprod,eq:sigmaprod,eq:x_distribution,eq:w_distribution,eq:Ex}, the variance of $j$-th element in $\mathbf{x}_{i+1}$ is
\begin{equation}
    \begin{aligned}
        \mathbb{\sigma}^2 (\mathbf{x}_{i+1}^j) 
        &= \mathbb{\sigma}^2 (\sum_{k=1}^{w_i} \mathbf{M}_i^{jk} \mathbf{x}_i^{k}) \\
        &= \sum_{k=1}^{w_i} \mathbb{\sigma}^2 (\mathbf{M}_i^{jk} \mathbf{x}_i^{k}) \\
        &= 
        \!\begin{multlined}[t]
        \sum_{k=1}^{w_i} \{ \mathbb{\sigma}^2 (\mathbf{M}_i^{jk}) \mathbb{\sigma}^2 (\mathbf{x}_i^{k})
        + \mathbb{\sigma}^2 (\mathbf{M}_i^{jk}) \mathbb{E} (\mathbf{x}_i^{k}) \\
        + \mathbb{\sigma}^2 (\mathbf{x}_i^{k}) \mathbb{E} (\mathbf{M}_i^{jk}) \} \\
        \end{multlined} \\
        &= \sum_{k=1}^{w_i}  \mathbb{\sigma}^2 (\mathbf{x}_i^{k}) \\
        &= {w_i}  \mathbb{\sigma}^2 (\mathbf{x}_i^{k}).
    \end{aligned}
\end{equation}
With the variances propagating in networks and $\mathbf{x}_1^j \sim{\mathcal{N}(0, 1)}$ in Eq.~\ref{eq:x_distribution}, the variance of $j$-th element in $L$-th layer is
\begin{equation}
    \mathbb{\sigma}^2 (\mathbf{x}_{L}^j) = \prod_{i=1}^{L} w_i
\end{equation}

According to Eq.~\ref{eq:entropy}, the entropy of each element $x_L^{j}$ of $L$-th MLP is 
\begin{equation}
    \begin{aligned}
        H(\mathbf{x}_L^{j}) 
        &\propto \log(\prod_{i=1}^{L} w_i), \\
        &= \sum_{i=1}^{L} \log(w_i).
    \end{aligned}
\end{equation}

After considering the width of the output feature vector, the normalized Gaussian entropy upper bound of the $L$-th feature map of MLP $f(\cdot)$ is
\begin{equation}
    H_f = w_{L+1}\sum_{i=1}^{L} \log(w_i).
\end{equation}

\section{Proof of Proposition~\ref{pro:average-width-of-mlp}} \label{sec:proof_compute_width}
Assume there is an MLP model $f_A(\cdot)$ that has $L$-layers with different width $w_i$, and the entropy of the MLP model is $H$. To define the ``average width'' of $f_A(\cdot)$,  we compare $f_A(\cdot)$ to a new MLP $f_B(\cdot)$. $f_B(\cdot)$ also has $L$-layers but with all layers sharing the same width $\bar{w}$. Suppose that the two networks have the same entropy \xuan{for each output neuron}, that is,
\begin{equation} \label{eq:proof_proportion_3}
    H_{f_a}=\sum_{i=1}^{L} \log(w_i), \quad
    H_{f_b} = L \cdot \log(\bar{w}).
\end{equation}

When the above equality holds true(i.e., $H_{f_a}=H_{f_b}$), \xuan{we can have the following equation,}
\begin{equation} \label{eq:proof_proportion_3}
    \sum_{i=1}^{L} \log(w_i) = L \cdot \log(\bar{w}).
\end{equation}

Therefore, we define $\bar{w}$ as the ``average width'' of $f_A(\cdot)$. Then we derive the definition of  average width of MLP in Proposition~\ref{pro:average-width-of-mlp} \xuan{as following,}
\begin{equation} \label{eq:proof_proportion_3}
    \bar{w} = \exp \left( \frac{1}{L} \sum_{i=1}^{L} \log w_i \right). \
\end{equation}

\section{SOTA DeepMAD Models} \label{sec:base_level_results}
We provide more SOTA DeepMAD models in
\cross{Table~\ref{tab:baselevel} and}
Figure~\ref{fig:full_sota}.
Especially, Deep-MAD achieves better performance on small and base level. DeepMAD-50M can achieve 83.9\% top-1 accuracy, which is even better than ConvNeXt-Base~\cite{liu2022convnet} with nearly only half of its scale.
DeepMAD-89M achieves 84.0\% top-1 accuracy at the ``base" scale, outperforming ConvNeXt-Base and Swin-Base~\cite{liu2021swin}, and achieves similar accuracy with SLaK-Base~\cite{liu2022more} with less computation cost and smaller model size.
On ``tiny" scale, DeepMAD-29M also achieves 82.8\% top-1 accuracy under 4.5G FLOPs and 29M Params. It is 1.5\% higher than Swin-Tiny with the same scale, and is 2.2x reduction in Params and 3.3x reduction in FLOPs compared to T2T-24~\cite{yuan2021tokens} with 0.2\% higher accuracy.
Therefore, we can find that building networks only with convolutional blocks can achieve better performance than those networks built with vision transformer blocks, which shows the potentiality of the convolutional blocks.

\section{DeepMAD Optimized for GPU Inference Throughput} \label{sec:throughput_results}

We optimize GPU inference throughput using DeepMAD. To measure the throughput, we use float32 precision (FP32) and increase the batch size for each model until no more images can be loaded in one mini-batch inference. The throughput is tested on NVIDIA V100 GPU with 16 GB Memory. ResNet building block is used as our design space.

To align with the throughput of ResNet-50 and Swin-Tiny on GPU, we first use DeepMAD to design networks of different Params and FLOPs. Then we test throughput for all models. Among these models, we choose two models labeled as DeepMAD-R50-GPU and DeepMAD-ST-GPU such that the two models are aligned with ResNet-50 and Swin-Tiny respectively. The top-1 accuracy on ImageNet-1k are reported in Table~\ref{tab:throughput}.



\begin{table}[h]
  \centering
  \resizebox{1.0\linewidth}{!}{
  \begin{tabular}{cccccc}
    \toprule
    Method & \makecell{Res.} & \#Param. & Throughput & FLOPs & Acc.(\%) \\
    \midrule
    ResNet-50 & 224 & 26\,M & 1245 & 25.6\,G & 77.4 \\
    DeepMAD-R50-GPU & 224 & 19\,M & 1171 & 3.0\,G & 80.0 \\
    \midrule
    Swin-Tiny & 224 & 29\,M & 750  & 4.5\,G & 81.3 \\
    DeepMAD-ST-GPU & 224 & 40\,M & 767 & 6.0\,G & 81.7 \\ 
    \bottomrule
  \end{tabular}
  }
  \caption{DeepMAD models optimized for throughput on GPU. `Res': image resolution.}
  \label{tab:throughput}
\end{table}

\section{Other Experiments Results} \label{sec:base_level_results}

We fine-tune the \textit{effectiveness} $\rho$ in Table~\ref{tab:finetue-rho}. Comparing to the results in the main text, we can achieve better accuracy when $\rho$ is fine-tuned.

\begin{table}[h]
  \centering
  \resizebox{1.0\linewidth}{!}{
  \begin{tabular}{cccccc}
    \toprule
    Method & $\rho$ & \makecell{Res.} & \#Param. & FLOPs & Acc.(\%) \\
    \midrule
    ResNet-18~\cite{He2015resnet} & 0.01 & 224 & 11.7\,M & 1.8\,G & 70.9 \\
    DeepMAD-R18 & 0.3 & 224 &11.7\,M & 1.8\,G  & 77.7 \\
    \rowcolor{light-gray} DeepMAD-R18 & 0.15 & 224 & 11.7\,M & 1.8\,G &  \textbf{78.2} \\
    \midrule
    ResNet-34~\cite{He2015resnet} & 0.02 & 224 & 21.8\,M & 3.6\,G & 74.4 \\
    DeepMAD-R34 & 0.3 & 224 & 21.8\,M & 3.6\,G & 79.7 \\
    \rowcolor{light-gray} DeepMAD-R34 & 0.15 & 224 & 21.8\,M & 3.6\,G &  \textbf{80.3} \\
    \midrule
    MobileNet-V2~\cite{howard2017mobilenets} & 0.9 & 224 & 3.5\,M & 320\,M & 72.0 \\
    DeepMAD-MB & 0.5 & 224 & 3.5\,M & 320\,M & 72.3 \\
    \rowcolor{light-gray} DeepMAD-MB & 1 & 224 & 3.5\,M & 320\,M &  \textbf{72.9} \\
    \bottomrule
  \end{tabular}
  }
  \caption{Fine-tuned $\rho$ in DeepMAD. `Res': image resolution.}
  \label{tab:finetue-rho}
\end{table}



\section{Complexity Comparison with NAS Methods} \label{sec:compare_to_nas_results}
DeepMAD is also compared with classical NAS works in complexity as well as accuracy on ImageNet-1K.
The classical NAS methods~\cite{zoph2018nasnet,cai2018proxylessnas,liu2018progressive,xie2018snas,real2019regularized} need to train a considerable number of networks and evaluate them the in searching phase, which is time-consuming and computing-consuming.
DeepMAD need not train any model in the search phase, and it just needs to solve the MP problem to obtain optimized network architectures in a few minutes.
As shown in Table~\ref{tab:comparetoNASworks}, DeepMAD takes only a few minutes to search for a network that can achieve better accuracy~(76.1\%) than other NAS methods.
It should be noted that Table~\ref{tab:comparetoNASworks} only consider the search time cost and does not consider the training time cost.
However, DeepMAD only needs one training process to produce a high-accuracy model with trained weights, while these baseline methods train multiple times.

\begin{table}[h]
  \centering
  \resizebox{1.0\linewidth}{!}{
  \begin{tabular}{ccccc}
    \toprule
    Method  & \#Param. & FLOPs & Acc.(\%) & \makecell{Search Cost\\(GPU hours)} \\
    \midrule
    NASNet-A~\cite{zoph2018nasnet} & 5.3\,M & 564\,M & 74.0 & 48,000 \\
    ProxylessNAS~\cite{cai2018proxylessnas} & 5.8\,M & 595\,M  & 76.0 & 200 \\
    PNAS~\cite{liu2018progressive} & 5.1\,M & 588\,M & 74.2 & 5,400\\
    SNAS~\cite{xie2018snas} & 4.3\,M & 522\,M & 72.7 & 36 \\
    AmoebaNet-A~\cite{real2019regularized} & 5.1\,M & 555\,M & 74.5 & 75,600 \\
    \midrule
    DeepMAD & 5.3\,M & 390\,M & 76.1 & \makecell{$<1$ \\(CPU hour)} \\
    \bottomrule
  \end{tabular}
}
  \caption{Complexity and accuracy comparison with NAS Methods on ImageNet-1K.}
  \label{tab:comparetoNASworks}
\end{table}



\section{Discussion on Architectures}
\begin{figure}[h]
  \centering
  \includegraphics[width=1.0\linewidth]{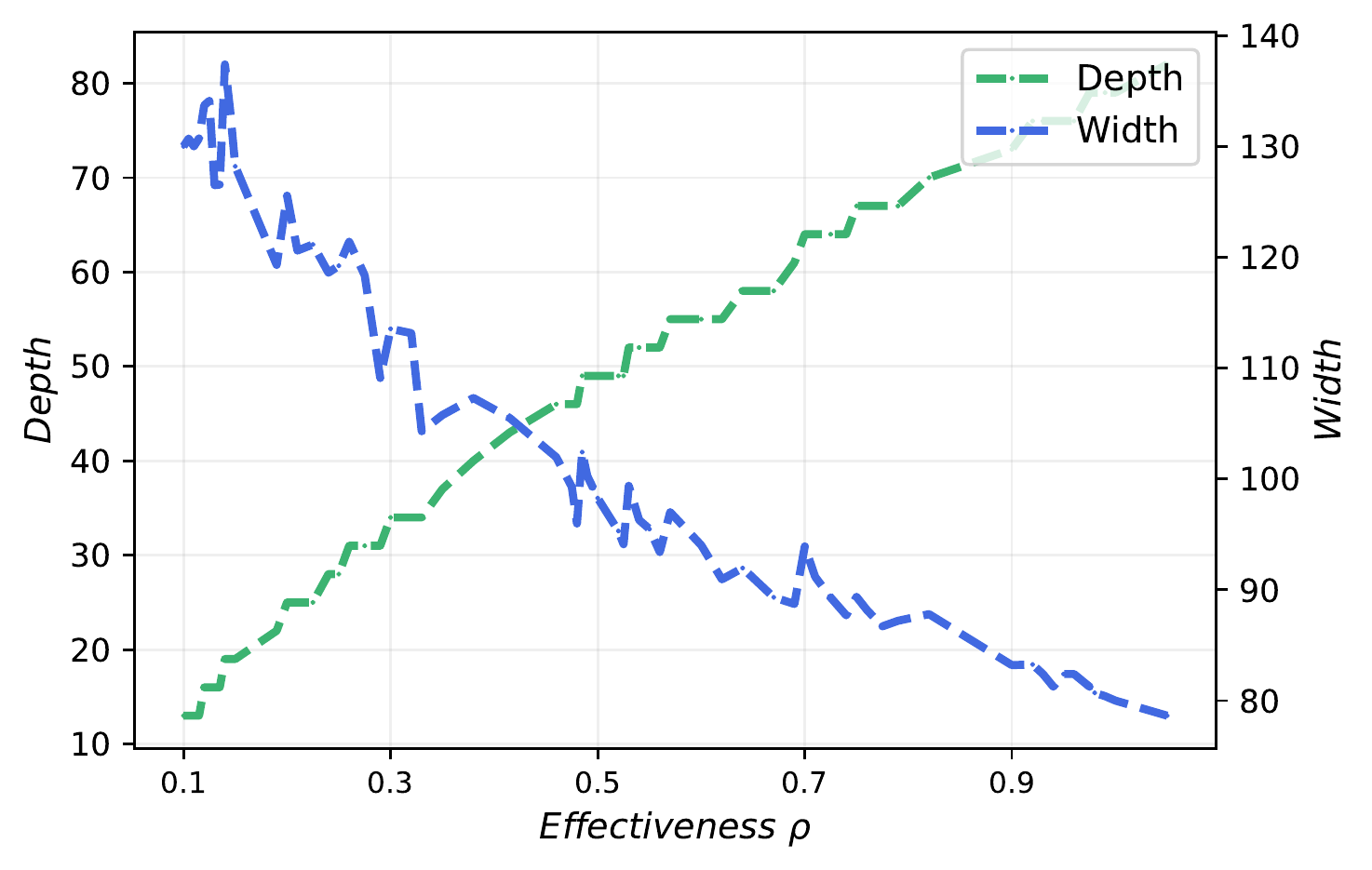}
  \caption{
  {\it Effectiveness} $\rho$ v.s. the depth and width of each generated network on CIFAR-100. The architectures shown in this figure are same as those shown in Figure~\ref{fig:rho_ablation_cifar_1}. The depth increases with $\rho$ monotonically while the width decreases at the same time.
  }
  \label{fig:rho_width_depth}
\end{figure}

Figure~\ref{fig:rho_width_depth} shows as \textit{effectiveness} $\rho$ increases, the depth of networks increases while the width decreases. As discussed in Section~\ref{section:Validations on CIFAR-100}, the model accuracy does not always increase with $\rho$. When $\rho$ is small, model accuracy increases as depth increases and width decreases. When $\rho$ is large, the opposite phenomenon occurs. The existence of an optimal \textit{effectiveness} means the existence of optimal depth and width of networks to reach the best accuracy.

The architectures of DeepMAD models are released along with the source codes.
Compared to ResNet families, DeepMAD suggests deeper and thinner structures. The final stage of DeepMAD networks is also deeper.
The width expansion after each downsampling layer is around 1.2-1.5, which smaller than 2 in ResNets.


\section{Experiment Settings}
The detailed training hyper-parameters for CIFAR-100 and ImageNet-1K datasets are reported in Table~\ref{tab:experiment-hyperparameters}.

\begin{table}[h]
  \centering
  \resizebox{1.0\linewidth}{!}{
  \begin{tabular}{lccccc}
    \toprule
    Hyper-parameter & CIFAR-100 & ImageNet-1K \\
    \midrule
    warm-up epoch & 5 & 20 \\
    cool-down epoch & 0 & 10 \\
    epochs & 1440 & 480 \\
    optimizer & SGD & SGD \\
    batchnorm momentum & 0.01 & 0.01 \\
    weight decay & 5e-4 & 5e-5 \\
    nesterov & True & True \\
    lr scheduler & cosine & cosine \\
    label smoothing & 0.1 & 0.1 \\
    mix up & 0.2 & 0.8 \\
    cut mix & 0 & 1.0 \\
    mixup switch prob & 0.5 & 0.5 \\
    crop pct & 0.875 & 0.95 \\
    reprob & 0.5 & 0.2 \\
    auto augmentation & auto~\cite{cubuk2018autoaugment} & rand-m9-mstd0.5 \\
    lr & 0.2 & 0.8 \\
    batch size & 512 & 2048 \\
    amp & False & True \\
    \bottomrule
  \end{tabular}
  }
  \caption{Experiment Settings.}
  \label{tab:experiment-hyperparameters}
\end{table}

\end{document}